\pdfoutput=1

\documentclass[11pt]{article}

\usepackage[preprint]{acl}
\usepackage{times}
\usepackage{latexsym}
\usepackage[T1]{fontenc}
\usepackage[utf8]{inputenc}
\usepackage{xcolor,colortbl}
\usepackage{microtype}
\usepackage{inconsolata}
\usepackage{graphicx}
\usepackage{booktabs}       
\usepackage{hyperref}
\usepackage{ulem}
\usepackage{enumitem}
\usepackage{footnote}

\definecolor{mycolumncolor}{RGB}{220, 240, 255}

\definecolor{tablecolor}{rgb}{0.8,0.8,0.8}

\newcommand\cut[1]{}

\newcommand{\squishlist}{
   \begin{list}{$\bullet$}
    { \setlength{\itemsep}{0pt}      \setlength{\parsep}{3pt}
      \setlength{\topsep}{3pt}       \setlength{\partopsep}{0pt}
      \setlength{\leftmargin}{1.5em} \setlength{\labelwidth}{1em}
      \setlength{\labelsep}{0.5em} } }

\newcommand{\squishlisttwo}{
   \begin{list}{$\bullet$}
    { \setlength{\itemsep}{0pt}    \setlength{\parsep}{0pt}
      \setlength{\topsep}{0pt}     \setlength{\partopsep}{0pt}
      \setlength{\leftmargin}{2em} \setlength{\labelwidth}{1.5em}
      \setlength{\labelsep}{0.5em} } }

\newcommand{\squishend}{
    \end{list}  }

{}
{}
{}

\newcommand{\blockcomment}[1]{}

\newcommand*\iftodonotes{\if@todonotes@disabled\expandafter\@secondoftwo\else\expandafter\@firstoftwo\fi}  
\makeatother

\title{Lugha-Llama: Adapting Large Language Models for African Languages} 

\makeatletter
\AtBeginDocument{%
\renewcommand{\sectionautorefname}{\S\@gobble}

\renewcommand{\subsectionautorefname}{\S\@gobble} 
\renewcommand{\subsubsectionautorefname}{\S\@gobble} 
}
\makeatother

\author{
    Happy Buzaaba$^{\dagger*}$ \; Alexander Wettig$^{\dagger*}$ \;
    David Ifeoluwa Adelani$^{\ddagger}$ \; Christiane Fellbaum$^{\dagger}$ \\
     \fontsize{11}{12}\selectfont{$^{\dagger}$Princeton University} \;
    \fontsize{11}{12}\selectfont{$^{\ddagger}$Mila, McGill University \& Canada CIFAR AI Chair} \\
}

\begin{document}
\maketitle

\renewcommand{\thefootnote}{\fnsymbol{footnote}}
\footnotetext[1]{The first two authors contributed equally. Correspondence to \texttt{\{happy.buzaaba@,awettig@cs.\}princeton.edu}.}

\renewcommand{\thefootnote}{\arabic{footnote}}
\setcounter{footnote}{0}

\begin{abstract}
Large language models (LLMs) have achieved impressive results in a wide range of natural language applications.
However, they often struggle to recognize low-resource languages, in particular African languages, which are not well represented in large training corpora.
In this paper, we consider how to adapt LLMs to low-resource African languages.
We find that combining curated data from African languages with high-quality English educational texts results in a training mix that substantially improves the model's performance on these languages.
On the challenging IrokoBench dataset, our models consistently achieve the best performance amongst similarly sized baselines, particularly on knowledge-intensive multiple-choice questions (AfriMMLU). Additionally, on the cross-lingual question answering benchmark AfriQA, our models outperform the base model by over 10\%.
To better understand the role of English data during training, we translate a subset of 200M tokens into Swahili language and perform an analysis which reveals that the content of these data is primarily responsible for the strong performance.  
We release our models\footnote{\url{https://huggingface.co/Lugha-Llama}} and data\footnote{\url{https://huggingface.co/datasets/princeton-nlp/fineweb_edu-swahili-translated}} to encourage future research on African languages.
\end{abstract}

\section{Introduction}
Large Language Models (LLMs) have
achieved remarkable progress in Natural Language
Processing and beyond \citep{DBLP:conf/nips/BrownMRSKDNSSAA20, chowdhery2023palm, touvron2023llama, chung2024scaling, DBLP:conf/emnlp/ZhangLB23}. Although LLMs perform well on a wide range of tasks in high-resource languages~\citep{Minaee2024LargeLM, Huang2024ASO}, their performance in low-resource languages, especially African languages, continues to lag behind \citep{DBLP:conf/africanlp/OjoO23, DBLP:journals/corr/abs-2302-09210}. Africa is home to approximately one third of the world's languages~\citep{ethnologue}, but these languages are underrepresented in commonly used pre-training datasets \citep{conneau-etal-2020-unsupervised}. As a result, LLMs are pre-trained on multilingual imbalanced datasets \citep{DBLP:conf/iclr/ChungGRTFNC23,Touvron2023Llama2O}, posing a significant challenge for multilingual models to recognize low-resource African languages. 

In a recent line of research, LLMs that were trained primarily on English data are adapted to African languages in an additional training phase on African language corpora \citep{uemura_afriinstruct}.
However, on challenging evaluations that require reasoning and expert knowledge, such as AfriMMLU  \citep{adelani2024irokobench},
these models lag far behind closed-source frontier models like GPT-4~\citep{openai2023gpt4}.

In this paper, we produce the African-centric {\tt Lugha-Llama}\footnote{\tt Lugha is the Kiswahili word for ``language''} by continuing to pre-train Llama-3.1-8B \citep{dubey2024llama} on 10B multilingual tokens.
Besides using a strong base model, we demonstrate the importance of curating the training data. Surprisingly, we find that adding high-quality English educational documents to the training data can further increase model performance in African language evaluations. Our model's overall performance on IrokoBench \citep{adelani2024irokobench} is the best of any current open-weight model.

Why is English data beneficial?
We analyze this by translating 200M tokens of educational English documents to Swahili using GPT-4o \citep{openai2023gpt4}.
The translated data perform substantially better than the English source data, suggesting that adding English is not necessary. However, it also outperforms existing high-quality Swahili corpora \citep{oladipo2023better}. This suggests that a consequence of data scarcity in low-resource languages is a gap in data quality compared to high-resource languages. It also raises the possibility of reducing this gap via large-scale machine translation.

We open-source the Lugha-Llama models and the 200M token corpus of educational documents translated to Swahili.

\section{Background} \label{sec:background}
\paragraph{Low-resource languages.} 
According to \cite{joshi-etal-2020-state} many African languages are categorized as low-resource, and there is a growing interest in NLP research to address challenges faced by such languages. A number of studies have focused on creating task-specific benchmark datasets like Named Entity Recognition \cite{adelani2021masakhaner}, Part-of-Speech tagging \cite{dione2023masakhapos} and Machine Translation \cite{adelani-etal-2022-thousand} to enable research on low-resource African languages. The most challenging of the African benchmarks is IrokoBench~\citep{adelani2024irokobench}---that focuses on mathematics reasoning, knowledge QA and natural language inference. In our work, we evaluate on AfriQA and IrokoBench showing progress on challenging benchmarks.

\paragraph{Multi-lingual language models.}
Multilingual models are a promising approach to work with low-resource languages.
Models such as mBERT~\cite{devlin2019bert}, XLM-R~\cite{conneau2019unsupervised} and mT5~\cite{xue2020mt5} jointly pre-train large models on over 100 languages, but only a handful of African languages are included.
Another approach is to use a small dataset to pre-train multilingual models from scratch as in \cite{ogueji-etal-2021-small,adebara2022serengeti,tonja2024inkubalm}, who release comparatively smaller models that in some cases match larger models pre-trained on much more data.
Several other studies have adapted multi-lingual models through continued pre-training~\citep{liu2021continual,lu2024llamax,Wu2024AdaptingLL, Lin2024MaLA500ML,alabi2022adapting,adelani-etal-2022-thousand} and instruction tuning \cite{muennighoff2022crosslingual, huang2023not, singh-etal-2024-aya}. 
The promise of continual pre-training is that some of the base model's capabilities seen with a high-resource language will transfer to a low-resource setting with comparatively little training compared to training from scratch.
Therefore, we build on the existing studies and focus on continual pre-training to develop an Africa-centric language model.

\begin{figure}[t]
    \includegraphics[width=0.9\linewidth]{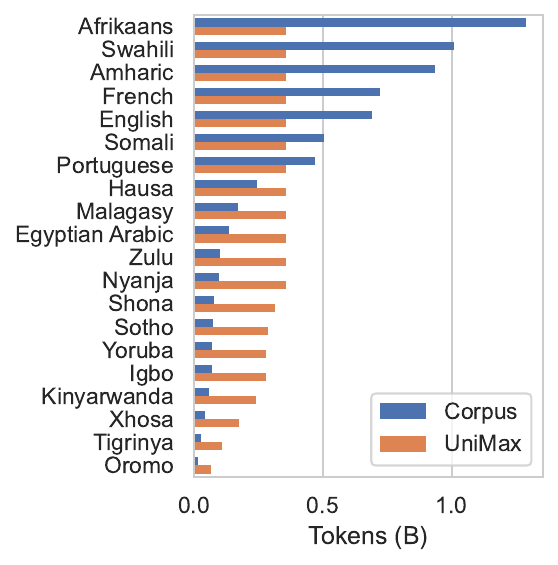}
    \caption{Tokens per language in the WURA corpus \citep{oladipo2023better} and the 6B-token training data for training using UniMax sampling \citep{chung2023unimax}.}
    \label{fig:wura_proportions}
\end{figure}

\paragraph{Data Quality.}
Data curation has emerged as an important part of language model development, as it can have a substantial impact of downstream task performance \citep{li2024datacomplm, albalak2024survey,oladipo2023better}.
Techniques include heuristic filters to remove internet artifacts \citep{raffel2020exploring, rae2021scaling, penedo2023refinedweb}, identifying duplicated documents \citep{lee2022deduplicating,abbas2023semdedup}, and training classifiers to identify documents based on some attribute, such as educational value \citep{wettig2024qurating, penedo2024finewebdatasetsdecantingweb}. 
While this focus on data quality has achieved strong results on English-language corpora, little has been explored about the relationship of data quality and data availability across languages, as well as the implications of this on mixing multi-lingual data sources.

\section{Experimental Setup}

\newcommand{\cc}[1]{{\cellcolor[HTML]{#1}}}

\begin{table*}[t]
    \centering
    \small
    
\resizebox{\textwidth}{!}{%
\setlength{\tabcolsep}{2pt}%
\setlength{\defaultaddspace}{5pt}%
\setlength{\aboverulesep}{0pt}%
\setlength{\belowrulesep}{0pt}%
\begin{tabular}{@{\hspace{2pt}}lcccccccccccccccccccc@{\hspace{2pt}}}
\toprule
 Model & Size & \textbf{eng} & \textbf{fra} & \textbf{amh} & \textbf{hau} & \textbf{ibo} & \textbf{kin} & \textbf{orm} & \textbf{sna} & \textbf{sot} & \textbf{swa} & \textbf{xho} & \textbf{yor} & \textbf{zul} & \textbf{\textit{ewe}} & \textbf{\textit{lin}} & \textbf{\textit{lug}} & \textbf{\textit{twi}} & \textbf{\textit{wol}} & \textbf{Avg}$^\dagger$ \\
\midrule
&& \multicolumn{19}{c}{\tt AfriMMLU} \\
\midrule
Llama-3.1 & 8B & \cc{70b08d} 61.4 & \cc{70b08d} 47.8 & \cc{bbd8c8} 30.6 & \cc{b7d5c5} 31.4 & \cc{bcd8c9} 30.4 & \cc{bad7c7} 31.0 & \cc{c3dcce} 29.2 & \cc{bdd9ca} 30.2 & \cc{bad7c7} 31.0 & \cc{a6ccb7} 34.6 & \cc{cfe2d7} 27.0 & \cc{bad7c7} 30.8 & \cc{bcd8c9} 30.4 & \cc{d4e5dc} 26.0 & \cc{aed1be} 33.0 & \cc{c1dbcc} 29.6 & \cc{d0e3d8} 26.8 & \cc{c7ded1} \bf 28.4 & \cc{bed9ca} 30.0 \\
\addlinespace
\bf Lugha-Llama & 8B & \cc{70b08d} 65.4 & \cc{70b08d} 51.2 & \cc{98c5ac} \bf 37.2 & \cc{a1cab4} 35.4 & \cc{a1cab3} 35.6 & \cc{a2cbb5} \bf 35.2 & \cc{abcfbb} 33.6 & \cc{9cc7b0} 36.4 & \cc{95c3aa} 37.8 & \cc{91c2a7} 38.4 & \cc{acd0bc} 33.4 & \cc{b3d3c1} 32.2 & \cc{a6ccb7} 34.6 & \cc{d5e5dc} 25.8 & \cc{aacfbb} 33.8 & \cc{c4dccf} 29.0 & \cc{d3e4db} 26.2 & \cc{d4e5dc} 26.0 & \cc{aed1bd} 33.2 \\
\bf $\;\;$-edu & 8B & \cc{70b08d} \bf 66.6 & \cc{70b08d} \bf 51.4 & \cc{99c5ad} 37.0 & \cc{93c3a9} \bf 38.0 & \cc{91c2a7} \bf 38.4 & \cc{a9ceba} 34.0 & \cc{a6ccb7} 34.6 & \cc{97c4ab} \bf 37.4 & \cc{93c3a9} \bf 38.0 & \cc{8dbfa4} \bf 39.2 & \cc{a9ceba} \bf 34.0 & \cc{a7cdb8} 34.4 & \cc{98c5ac} \bf 37.2 & \cc{d0e3d8} \bf 26.8 & \cc{a7cdb8} \bf 34.4 & \cc{bad7c7} \bf 30.8 & \cc{cadfd3} \bf 28.0 & \cc{cee1d6} 27.2 & \cc{a7cdb8} \bf 34.3 \\
\bf $\;\;$-math & 8B & \cc{70b08d} 65.8 & \cc{70b08d} \bf 51.4 & \cc{99c5ad} 37.0 & \cc{9ac6ae} 36.8 & \cc{9ec8b1} 36.0 & \cc{aacfbb} 33.8 & \cc{9cc7b0} \bf 36.4 & \cc{a5ccb6} 34.8 & \cc{a3cbb5} 35.0 & \cc{8ec0a4} 39.0 & \cc{bfdacb} 30.0 & \cc{9dc8b0} \bf 36.2 & \cc{aacfbb} 33.8 & \cc{deeae3} 24.2 & \cc{abcfbb} 33.6 & \cc{c8ded2} 28.2 & \cc{cfe2d7} 27.0 & \cc{d6e6dd} 25.6 & \cc{afd1be} 33.0 \\

\addlinespace
aya-101 & 13B &  \cc{7bb695} 42.6 & \cc{88bc9f} 40.2 & \cc{a1cab4} 35.4 & \cc{abcfbb} 33.6 & \cc{9cc7b0} 36.4 & \cc{b1d2c0} 32.6 & \cc{bdd9ca} 30.2 & \cc{b8d6c5} 31.2 & \cc{b0d2bf} 32.8 & \cc{a3cbb5} 35.0 & \cc{b2d3c0} 32.4 & \cc{abcfbb} 33.6 & \cc{a6ccb7} 34.6 & \cc{d1e3d9} 26.6 & \cc{bfdacb} 30.0 & \cc{c0dacc} 29.8 & \cc{cfe2d7} 27.0 & \cc{dbe8e1} 24.8 & \cc{b6d5c4} 31.6 \\
LLaMAX3 & 8B & \cc{70b08d} 55.4 & \cc{73b28f} 44.0 & \cc{bfdacb} 30.0 & \cc{add0bd} 33.2 & \cc{b0d2bf} 32.8 & \cc{c5ddd0} 28.8 & \cc{bfdacb} 30.0 & \cc{b2d3c0} 32.4 & \cc{c5ddd0} 28.8 & \cc{a2cbb5} 35.2 & \cc{c2dbcd} 29.4 & \cc{c0dacc} 29.8 & \cc{bad7c7} 31.0 & \cc{e8efeb} 22.4 & \cc{b0d2bf} 32.8 & \cc{cfe2d7} 27.0 & \cc{d7e6de} 25.4 & \cc{d6e6dd} 25.6 & \cc{c0dacc} 29.7 \\
AfroLlama-V1 & 8B & \cc{9cc7b0} 36.4 & \cc{e5ede8} 23.0 & \cc{dce9e1} 24.6 & \cc{bcd8c9} 30.4 & \cc{d2e4da} 26.4 & \cc{dfebe4} 24.0 & \cc{e2ece6} 23.4 & \cc{dae8e0} 25.0 & \cc{d9e7df} 25.2 & \cc{c7ded1} 28.4 & \cc{d2e4da} 26.4 & \cc{d4e5dc} 26.0 & \cc{cfe2d7} 27.0 & \cc{d7e6de} 25.4 & \cc{dfebe4} 24.0 & \cc{ebf1ed} 21.8 & \cc{d9e7df} 25.2 & \cc{e8efeb} 22.4 & \cc{d8e7de} 25.3 \\
AfriInstruct & 7B & \cc{77b493} 43.2 & \cc{abcfbb} 33.6 & \cc{edf2ef} 21.4 & \cc{d4e5dc} 26.0 & \cc{dfebe4} 24.0 & \cc{d0e3d8} 26.8 & \cc{cee1d6} 27.2 & \cc{c8ded2} 28.2 & \cc{cbe0d4} 27.8 & \cc{bdd9ca} 30.2 & \cc{bad7c7} 30.8 & \cc{d3e4db} 26.2 & \cc{c8ded2} 28.2 & \cc{d4e5dc} 26.0 & \cc{cfe2d7} 27.0 & \cc{dfebe4} 24.0 & \cc{cbe0d4} 27.8 & \cc{d6e6dd} 25.6 & \cc{d1e3d9} 26.7 \\
InkubaLM & 0.4B & \cc{dfebe4} 24.0 & \cc{e4ede8} 23.2 & \cc{d7e6de} 25.4 & \cc{d5e5dc} 25.8 & \cc{d9e7df} 25.2 & \cc{e1ece6} 23.6 & \cc{cfe2d7} 27.0 & \cc{ebf1ed} 21.8 & \cc{d4e5dc} 26.0 & \cc{d7e6de} 25.4 & \cc{dae8e0} 25.0 & \cc{e1ece6} 23.6 & \cc{deeae3} 24.2 & \cc{deeae3} 24.2 & \cc{e1ece6} 23.6 & \cc{e0ebe5} 23.8 & \cc{d6e6dd} 25.6 & \cc{d0e3d8} 26.8 & \cc{dbe8e1} 24.8 \\
\midrule
&& \multicolumn{19}{c}{\tt AfriMGSM} \\
\midrule
Llama-3.1 & 8B & \cc{70b08d} \bf 16.0 & \cc{70b08d} \bf 12.4 & \cc{c9dfd2} 2.8 & \cc{a9ceba} 5.2 & \cc{c3dcce} 3.2 & \cc{aed1be} 4.8 & \cc{c9dfd2} 2.8 & \cc{a4cbb6} \bf 5.6 & \cc{bed9ca} 3.6 & \cc{7ab595} 8.8 & \cc{94c3a9} \bf 6.8 & \cc{a9ceba} 5.2 & \cc{c3dcce} 3.2 & \cc{c3dcce} \bf 3.2 & \cc{b4d4c2} 4.4 & \cc{aed1be} 4.8 & \cc{c9dfd2} 2.8 & \cc{b9d6c6} \bf 4.0 & \cc{b3d3c2} 4.5 \\
\addlinespace
\bf Lugha-Llama & 8B & \cc{70b08d} 11.2 & \cc{70b08d} 11.2 & \cc{a4cbb6} \bf 5.6 & \cc{8fc0a5} 7.2 & \cc{a4cbb6} \bf 5.6 & \cc{a9ceba} 5.2 & \cc{cee1d6} 2.4 & \cc{aed1be} 4.8 & \cc{b4d4c2} 4.4 & \cc{70b08d} 9.6 & \cc{a9ceba} 5.2 & \cc{8fc0a5} 7.2 & \cc{a4cbb6} 5.6 & \cc{c3dcce} \bf 3.2 & \cc{b4d4c2} 4.4 & \cc{9fc9b1} 6.0 & \cc{cee1d6} 2.4 & \cc{cee1d6} 2.4 & \cc{abcfbb} 5.1 \\
\bf $\;\;$-edu & 8B & \cc{70b08d} 13.2 & \cc{75b391} 9.2 & \cc{b4d4c2} 4.4 & \cc{8abea1} 7.6 & \cc{bed9ca} 3.6 & \cc{aed1be} 4.8 & \cc{aed1be} \bf 4.8 & \cc{b4d4c2} 4.4 & \cc{a9ceba} \bf 5.2 & \cc{75b391} 9.2 & \cc{b4d4c2} 4.4 & \cc{9fc9b1} 6.0 & \cc{8fc0a5} \bf 7.2 & \cc{cee1d6} 2.4 & \cc{aed1be} \bf 4.8 & \cc{b4d4c2} 4.4 & \cc{c9dfd2} 2.8 & \cc{c3dcce} 3.2 & \cc{acd0bc} 5.0 \\
\bf $\;\;$-math & 8B & \cc{70b08d} 15.2 & \cc{70b08d} 11.6 & \cc{a4cbb6} \bf 5.6 & \cc{70b08d} \bf 10.8 & \cc{a9ceba} 5.2 & \cc{9fc9b1} \bf 6.0 & \cc{b9d6c6} 4.0 & \cc{a4cbb6} \bf 5.6 & \cc{a9ceba} \bf 5.2 & \cc{70b08d} \bf 11.2 & \cc{9ac6ae} 6.4 & \cc{85bb9d} \bf 8.0 & \cc{9fc9b1} 6.0 & \cc{c9dfd2} 2.8 & \cc{aed1be} \bf 4.8 & \cc{9ac6ae} \bf 6.4 & \cc{bed9ca} \bf 3.6 & \cc{c3dcce} 3.2 & \cc{a0c9b2} \bf 5.9 \\
\addlinespace
aya-101 & 13B & \cc{8abea1} 7.6 & \cc{94c3a9} 6.8 & \cc{a9ceba} 5.2 & \cc{9fc9b1} 6.0 & \cc{cee1d6} 2.4 & \cc{c9dfd2} 2.8 & \cc{d3e4da} 2.0 & \cc{c3dcce} 3.2 & \cc{c9dfd2} 2.8 & \cc{b4d4c2} 4.4 & \cc{c9dfd2} 2.8 & \cc{bed9ca} 3.6 & \cc{c3dcce} 3.2 & \cc{d8e7df} 1.6 & \cc{d8e7df} 1.6 & \cc{cee1d6} 2.4 & \cc{cee1d6} 2.4 & \cc{ddeae3} 1.2 & \cc{c6ddd1} 3.0 \\
LLaMAX3 & 8B & \cc{b9d6c6} 4.0 & \cc{c3dcce} 3.2 & \cc{ddeae3} 1.2 & \cc{ddeae3} 1.2 & \cc{d8e7df} 1.6 & \cc{d8e7df} 1.6 & \cc{e3ece7} 0.8 & \cc{d8e7df} 1.6 & \cc{d3e4da} 2.0 & \cc{c9dfd2} 2.8 & \cc{e3ece7} 0.8 & \cc{ddeae3} 1.2 & \cc{cee1d6} 2.4 & \cc{e3ece7} 0.8 & \cc{d3e4da} 2.0 & \cc{d3e4da} 2.0 & \cc{e3ece7} 0.8 & \cc{bed9ca} 3.6 & \cc{d7e6de} 1.7 \\
AfroLlama-V1 & 8B & \cc{c3dcce} 3.2 & \cc{b9d6c6} 4.0 & \cc{e3ece7} 0.8 & \cc{bed9ca} 3.6 & \cc{e3ece7} 0.8 & \cc{d8e7df} 1.6 & \cc{edf2ef} 0.0 & \cc{cee1d6} 2.4 & \cc{e3ece7} 0.8 & \cc{b9d6c6} 4.0 & \cc{cee1d6} 2.4 & \cc{c3dcce} 3.2 & \cc{c9dfd2} 2.8 & \cc{e3ece7} 0.8 & \cc{cee1d6} 2.4 & \cc{cee1d6} 2.4 & \cc{e8efeb} 0.4 & \cc{e3ece7} 0.8 & \cc{d5e5dc} 1.8 \\
AfriInstruct & 7B & \cc{aed1be} 4.8 & \cc{c3dcce} 3.2 & \cc{d8e7df} 1.6 & \cc{c9dfd2} 2.8 & \cc{d8e7df} 1.6 & \cc{d8e7df} 1.6 & \cc{ddeae3} 1.2 & \cc{e3ece7} 0.8 & \cc{d8e7df} 1.6 & \cc{bed9ca} 3.6 & \cc{c9dfd2} 2.8 & \cc{d8e7df} 1.6 & \cc{ddeae3} 1.2 & \cc{ddeae3} 1.2 & \cc{ddeae3} 1.2 & \cc{e8efeb} 0.4 & \cc{e8efeb} 0.4 & \cc{ddeae3} 1.2 & \cc{d9e7df} 1.6 \\
InkubaLM & 0.4B & \cc{d8e7df} 1.6 & \cc{e8efeb} 0.4 & \cc{edf2ef} 0.0 & \cc{cee1d6} 2.4 & \cc{e8efeb} 0.4 & \cc{d8e7df} 1.6 & \cc{e3ece7} 0.8 & \cc{e3ece7} 0.8 & \cc{e8efeb} 0.4 & \cc{e3ece7} 0.8 & \cc{e3ece7} 0.8 & \cc{edf2ef} 0.0 & \cc{e3ece7} 0.8 & \cc{e8efeb} 0.4 & \cc{edf2ef} 0.0 & \cc{d3e4da} 2.0 & \cc{e8efeb} 0.4 & \cc{edf2ef} 0.0 & \cc{e4ede8} 0.7 \\

\midrule
&& \multicolumn{19}{c}{\tt AfriXNLI} \\
\midrule
Llama-3.1 & 8B & \cc{70b08d} 50.8 & \cc{70b08d} \bf 51.7 & \cc{c6ddd0} 35.0 & \cc{bdd9c9} 35.8 & \cc{d3e4db} 33.7 & \cc{bdd9c9} 35.8 & \cc{afd1be} 37.2 & \cc{bfdacb} 35.7 & \cc{d5e5dc} 33.5 & \cc{a5ccb6} 38.2 & \cc{e5ede8} 32.0 & \cc{c2dbcd} 35.3 & \cc{d7e6de} 33.3 & \cc{cbe0d4} 34.5 & \cc{e5ede8} 32.0 & \cc{d8e7df} 33.2 & \cc{c2dbcd} \bf 35.3 & \cc{d2e3d9} 33.8 & \cc{c9dfd3} 34.6 \\
\addlinespace
\bf Lugha-Llama & 8B & \cc{70b08d} 50.5 & \cc{70b08d} 50.8 & \cc{8fc0a5} \bf 40.3 & \cc{86bc9e} 41.2 & \cc{a7cdb8} 38.0 & \cc{9dc7b0} \bf 39.0 & \cc{a0c9b3} 38.7 & \cc{94c3a9} 39.8 & \cc{85bb9d} \bf 41.3 & \cc{70b08d} 43.5 & \cc{83ba9c} 41.5 & \cc{96c4ab} 39.7 & \cc{a1cab4} 38.5 & \cc{d8e7df} 33.2 & \cc{d3e4db} 33.7 & \cc{d7e6de} 33.3 & \cc{dae8e0} 33.0 & \cc{dce9e1} 32.8 & \cc{a7cdb8} 38.0 \\
\bf $\;\;$-edu & 8B & \cc{70b08d} \bf 51.3 & \cc{70b08d} \bf 51.7 & \cc{b1d2c0} 37.0 & \cc{83ba9c} 41.5 & \cc{9dc7b0} 39.0 & \cc{accfbc} 37.5 & \cc{9bc6ae} \bf 39.2 & \cc{9dc7b0} 39.0 & \cc{8dbfa4} 40.5 & \cc{70b08d} \bf 43.8 & \cc{88bda0} 41.0 & \cc{99c6ad} 39.3 & \cc{90c1a6} 40.2 & \cc{d5e5dc} 33.5 & \cc{cee2d7} \bf 34.2 & \cc{d8e7df} 33.2 & \cc{d2e3d9} 33.8 & \cc{d7e6de} 33.3 & \cc{a8cdb9} 37.9 \\
\bf $\;\;$-math & 8B & \cc{70b08d} 50.2 & \cc{70b08d} 51.5 & \cc{96c4ab} 39.7 & \cc{83ba9c} 41.5 & \cc{8fc0a5} \bf 40.3 & \cc{a0c9b3} 38.7 & \cc{9ec8b1} 38.8 & \cc{b3d3c1} 36.8 & \cc{92c2a8} 40.0 & \cc{71b18e} 43.2 & \cc{81b99a} \bf 41.7 & \cc{8fc0a5} 40.3 & \cc{85bb9d} \bf 41.3 & \cc{d5e5dc} 33.5 & \cc{d5e5dc} 33.5 & \cc{d7e6de} 33.3 & \cc{d7e6de} 33.3 & \cc{d8e7df} 33.2 & \cc{a6ccb7} \bf 38.1 \\
\addlinespace
aya-101 & 13B & \cc{85bb9d} 41.3 & \cc{accfbc} 37.5 & \cc{a1cab4} 38.5 & \cc{bfdacb} 35.7 & \cc{bad7c7} 36.2 & \cc{c6ddd0} 35.0 & \cc{d0e3d8} 34.0 & \cc{94c3a9} 39.8 & \cc{c6ddd0} 35.0 & \cc{a0c9b3} 38.7 & \cc{b8d6c5} 36.3 & \cc{afd1be} 37.2 & \cc{bbd8c8} 36.0 & \cc{bdd9c9} 35.8 & \cc{deeae3} 32.7 & \cc{b6d5c4} \bf 36.5 & \cc{d5e5dc} 33.5 & \cc{d3e4db} 33.7 & \cc{bcd8c9} 35.9 \\
LLaMAX3 & 8B & \cc{70b08d} 47.0 & \cc{70b08d} 48.7 & \cc{c7ded1} 34.8 & \cc{81b99a} \bf 41.7 & \cc{d5e5dc} 33.5 & \cc{c6ddd0} 35.0 & \cc{aed1bd} 37.3 & \cc{8fc0a5} \bf 40.3 & \cc{c9dfd3} 34.7 & \cc{70b08d} \bf 43.8 & \cc{a8ceb9} 37.8 & \cc{bbd8c8} 36.0 & \cc{c0dacc} 35.5 & \cc{d5e5dc} 33.5 & \cc{dae8e0} 33.0 & \cc{cde1d6} 34.3 & \cc{dae8e0} 33.0 & \cc{cbe0d4} \bf 34.5 & \cc{bad7c7} 36.2 \\
AfroLlama-V1 & 8B & \cc{70b08d} 44.0 & \cc{9bc6ae} 39.2 & \cc{d2e3d9} 33.8 & \cc{8bbea2} 40.7 & \cc{e8efeb} 31.7 & \cc{c9dfd3} 34.7 & \cc{bdd9c9} 35.8 & \cc{ecf1ee} 31.3 & \cc{dce9e1} 32.8 & \cc{7cb696} 42.2 & \cc{8bbea2} 40.7 & \cc{89bda1} \bf 40.8 & \cc{aacfbb} 37.7 & \cc{e6eee9} 31.8 & \cc{d3e4db} 33.7 & \cc{dce9e1} 32.8 & \cc{cee2d7} 34.2 & \cc{d0e3d8} 34.0 & \cc{c0dacc} 35.5 \\
AfriInstruct & 7B & \cc{70b08d} 51.0 & \cc{70b08d} 49.7 & \cc{cee2d7} 34.2 & \cc{8bbea2} 40.7 & \cc{afd1be} 37.2 & \cc{c0dacc} 35.5 & \cc{bdd9c9} 35.8 & \cc{9ec8b1} 38.8 & \cc{bad7c7} 36.2 & \cc{aed1bd} 37.3 & \cc{96c4ab} 39.7 & \cc{8dbfa4} 40.5 & \cc{a5ccb6} 38.2 & \cc{bbd8c8} \bf 36.0 & \cc{deeae3} 32.7 & \cc{e6eee9} 31.8 & \cc{cde1d6} 34.3 & \cc{dae8e0} 33.0 & \cc{b8d6c5} 36.4 \\
InkubaLM & 0.4B & \cc{edf2ef} 31.2 & \cc{dce9e1} 32.8 & \cc{d8e7df} 33.2 & \cc{c6ddd0} 35.0 & \cc{c4dccf} 35.2 & \cc{d8e7df} 33.2 & \cc{cee2d7} 34.2 & \cc{d8e7df} 33.2 & \cc{deeae3} 32.7 & \cc{e1ece6} 32.3 & \cc{cbe0d4} 34.5 & \cc{d0e3d8} 34.0 & \cc{d2e3d9} 33.8 & \cc{dae8e0} 33.0 & \cc{d3e4db} 33.7 & \cc{dce9e1} 32.8 & \cc{cbe0d4} 34.5 & \cc{d3e4db} 33.7 & \cc{d3e4db} 33.7 \\

\bottomrule
\end{tabular}
}
    \caption{Results of Lugha-Llama models and baselines on IrokoBench \citep{adelani2024irokobench}. Languages in \textit{italic} are not present in the continual pre-training data. $^\dagger$: We exclude the high-resource languages English (eng) and French (fra) from the average, as they would otherwise skew the results due to strong English base models.}
    \label{tab:main_results}
    \vspace{-1em}
\end{table*}

In our main experiments, we adapt a Llama-3.1-8B model to African Languages by training on 10B tokens with three different data mixtures.
Note that the base model was pre-trained on 15T tokens, of which 8\% are non-English \citet{dubey2024llama}.
\begin{itemize}[topsep=3pt,parsep=1pt,partopsep=1pt,leftmargin=1em]
    \item \textbf{(Lugha-Llama)} We sample 10B tokens from the WURA corpus \citep{oladipo2023better}, which is comprised of sixteen African languages and four high-resource languages commonly spoken on the African continent, namely English, French, Arabic, and Portuguese. The corpus was collected by inspecting and cleaning mC4 \citep{xue2020mt5} and crawling African websites.
    The corpus also includes 3 languages with non-latin scripts, namely Amharic, Arabic and Tigirinya.

    \item \textbf{(Lugha-Llama-edu)} We consider the effect of adding high-quality English educational documents to the WURA corpus. We combine 6B tokens from WURA with 4B tokens from FineWeb-Edu \citep{penedo2024finewebdatasetsdecantingweb}---a dataset obtained by prompting LLMs to score the educational content of web pages.

    \item \textbf{(Lugha-Llama-math)} To boost the mathematical reasoning abilities of African language models, we replace FineWeb-Edu with the OpenWebMath dataset \citep{paster2024openwebmath}, which contains documents with mathematical content identified via heuristic rules.
\end{itemize}
We sample from WURA using UniMax sampling \citep{chung2023unimax}, which attempts to sample as uniformly as possible across languages while limiting the number of times data is repeated. We upsample rare languages by at most four  epochs, since this has been found to incur no discernible degradation during model training  \citep{muennighoff2023scaling}. 
In \autoref{fig:wura_proportions}, we show the language proportions in the WURA corpus, as well as the training distribution with UniMax.

\paragraph{Training.} We initialize the models with Llama-3.1-8B \citep{dubey2024llama} and train with a batch size of 512 sequences containing 8192 tokens each. We train for 2400 steps, totalling 10B tokens, with a learning rate of $10^{-5}$ with a cosine learning rate schedule with 240 steps linear warmup, and decaying to $10^{-6}$.
We disable attention across document boundaries within a sequence, following the strategy used by Llama-3.1-8B during pre-training. 

\paragraph{Evaluation.} 
We make use of the EleutherAI LM Evaluation Harness~\cite{biderman2024lessons} to evaluate on the IrokoBench \citep{adelani2024irokobench}, a human translated benchmark for 16 typologically diverse low-resource African languages. This benchmark covers natural language inference (AfriXNLI), mathematical reasoning (AfriMGSM), and multi-choice knowledge-based question answering (AfriMMLU), which are derived from subsets of XNLI, MMLU, and MGSM respectively \citep{conneau-etal-2018-xnli, hendrycks2021measuring, shi2022language}. 
We use few-shot prompting and follow \citet{adelani2024irokobench} by setting the number of in-context demonstrations to eight for AfriMGSM and GSM8K, and 5 for all other tasks. We report the results with the language-specific prompt template for AfriXNLI, and the English templates for AfriMMLU and AfriMGSM, as these tasks do not provide a native language prompt template. Separately, we evaluate on AfriQA \cite{ogundepo-etal-2023-cross}, a cross-lingual open-retrieval question answering benchmark for African languages.

\paragraph{Baseline models.} 
We compare Lugha-Llama models to six open-weight LLMs:
Aya-101~\citep{ustun2024aya},
InkubaLM-0.4B~\citep{tonja2024inkubalm}, Llama-3.1-8B~\citep{dubey2024llama}, AfroLlama-v1 (8B)\footnote{\url{https://huggingface.co/Jacaranda/AfroLlama_V1}}, LLaMaX3-8B~\citep{lu2024llamax}, and
AfriInstruct~\citep{uemura_afriinstruct}. 
We report more details about the provenance of these models in \autoref{app:baselines}.

\begin{table}[h]
    \centering
    \small
    \resizebox{\linewidth}{!}{
    \begin{tabular}{lccc}
        \toprule
        \tt Model & Size & \tt Average Score \\
        \midrule
        Llama-3.1 & 8B  & 20.1 \\
        \midrule
        Lugha-Llama & 8B &  \bf{34.2} \\
        Lugha-Llama-Edu & 8B & \bf{30.3} \\
        Lugha-Llama-Math & 8B & \bf{37.7} \\
        \midrule
        AfroLlama-V1 & 8B & 19.0 \\
        AfriInstruct & 7B & 21.9 \\
        \bottomrule
        \end{tabular}
    }
    \caption{Results of Lugha-Llama models compared to base Llama and similarly sized Africa centric language models on AfriQA \cite{ogundepo-etal-2023-cross}.}
    \label{tab:Afriqa-results}
\end{table}

\begin{table}[t]
    \centering  
    \resizebox{\linewidth}{!}{%
    \setlength{\tabcolsep}{6pt}%
    \setlength{\defaultaddspace}{4pt}%
    \setlength{\aboverulesep}{0pt}%
    \setlength{\belowrulesep}{0pt}%
    \begin{tabular}{lccc}    
    \toprule
    & \multicolumn{3}{c}{\tt AfriMMLU} \\
    \cmidrule(lr){2-4}
     Training Data & \bf eng & \bf swa & \bf Avg$^\dagger$ \\
     \midrule
    100\% WURA$_{swa}$ & 64.4 & 41.0 & 30.6 \\
    \phantom{0}60\% WURA$_{swa}$ + 40\% FW-Edu & \bf 66.6 & 42.6 & \bf 31.6 \\
    \phantom{0}60\% WURA$_{swa}$ + 40\% FW-Edu$_{swa}$ & 65.4 & \bf 46.0 & 31.5 \\
    \phantom{060\% WURAWU!} 100\% FW-Edu$_{swa}$ & 66.2 & 43.8 & 31.5 \\
    \bottomrule
    \end{tabular}
    }
    \caption{We translate 200M-tokens from FineWeb-Edu to Swahili to disentangle the effects of semantic content from source language during continued pre-training.}
    \label{tab:swahili}
    \vspace{-1em}
\end{table}

\section{Results}
Our main results are shown in \autoref{tab:main_results}. Additional results and ablations can be found in \autoref{app:results}.

\label{sec:results}
\paragraph{Lugha-Llamas achieve strong results.} 
In \autoref{tab:main_results} we compare the three {Lugha-llama} models to the baselines on three tasks in IrokoBench. 
When considering the average scores over low-resource African languages, all three models perform better than all baselines across AfriMMLU, AfriMGSM, and AfriXNLI. The adaptation to African languages increases AfriMMLU scores by up to 8 percentage points compared to \mbox{Llama-3.1-8B}, with the largest improvements in Igbo (ibo).
However, the performance is similar to the Llama-3.1-8B baseline on languages not present in WURA. The results in \autoref{tab:Afriqa-results}
show that our Lugha-Llama models significantly outperform the base model by more than 10\% and consistently outperforms similarly sized Africa-centric language models on the cross-lingual question answering benchmark (AfriQA).

\paragraph{Addition of English data.}
We observe 
that including FineWeb-Edu data boosts the performance in AfriMMLU, and including data from OpenWebMath improves performance in AfriMGSM.
This is surprising, since these models are trained on 40\% less tokens from African languages.
However, it means that performance may still be improved by training on the leftover data, 
which is especially appealing in a low-resource settings.

\section{Case Study on Swahili Data Quality}
We have adapted a language model that was primarily pre-trained on English data
and saw improved performance on AfriMMLU when FineWeb-edu is part of the training data.
Such cross-lingual generalization in language models has been observed and studied by many works~\citep[inter alia]{artetxe-etal-2020-cross, deshpande-etal-2022-bert, ye2023language}.

We study the question whether it is \textit{necessary} for the FineWeb-Edu data to be in English to achieve these performance gains.
For example, given the English nature of the base model, it may prevent catastrophic forgetting, a common problem in continual learning \citep{wang2023comprehensive}, or better help the model integrate African languages better with its pre-trained representations.

We take a random sample of FineWeb-Edu and translate 130M English tokens into 190M Swahili tokens by prompting GPT-4o with individual documents.
We choose Swahili since we find that it is well supported by GPT-4o and is reported to have a better translation quality among many African languages \cite{costa2022no, robinson-etal-2023-chatgpt}.

We repeat experiments similar to our main results but train only for 1B tokens on Swahili or English data. 
\autoref{tab:swahili} shows that combining WURA with the translated educational data substantially outperforms the other data mixes. This suggests that the content of the FineWeb-Edu data is more critical to the performance than its English nature, and that a gap remains in terms of the pre-training data quality between English and low-resource languages such as Swahili. 

\section{Conclusion}
We introduce three Lugha-Llama models based on continued pre-training, which achieve best-in-class performance on the challenging IrokoBench tasks.
We demonstrate that combining African language pre-training data with educational English documents can improve downstream performance.
Our case study suggests that there is still a gap in data quality between English and low-resource languages, but raises the possibility of closing this gap via large-scale machine translation.

\section*{Limitations}
We discuss the limitations of our study of adapting language models to African languages.

\paragraph{Narrow evaluation.}
Our conclusions about the data mixture are sensitive to the choice of evaluations. We rely heavily on AfriMMLU, as MMLU is known to be an excellent signal for model quality. However, in a multilingual setting, a translated dataset like MMLU suffer from cultural bias  which introduces evaluation challenges and limit its practical effectiveness as a global benchmark \cite{singh2024global}. In addition, we notice that many questions look similar across languages as they use mathematical symbols and entity names that are largely unchanged in the languages. It is possible that the English FineWeb-edu data helps model make educated guesses based on surface cues in these cases.
Furthermore, some MMLU subjects are skewed towards Western and US-centric knowledge \citep{gema2024mmlu}. African-centric LLMs should also be developed and evaluated based on African-centric knowledge benchmarks.

\paragraph{English prompts.}
IrokoBench only provides English task prompts for AfriMMLU and AfriMGSM, meaning that part of the evaluation input is always in English. While this might also contribute towards the usefulness of additional English data, our case study in Swahili suggests that this should not be a major factor, since continued pre-training with Swahili translations performs the best.

\paragraph{Limited coverage of African languages.}
We only train on the WURA corpus, which covers 16 low-resource African languages. This ignores the great linguistic diversity on the African continent, and future work is necessary to extend the coverage of language models.

\paragraph{Computational cost.} Each model training run uses a considerable pre-training budget, requiring a total of 355 hours on Nvidia H100 GPUs. The resulting 8B-parameter model is also too expensive to run on most current consumer hardware.

\section*{Acknowledgment}

We thank Tianyu Gao, Sewon Min, and Sanjeev Arora for useful discussions. The authors gratefully acknowledges financial supportfrom Princeton Laboratory for Artificial Intelligence and  Princeton Language and Intelligence for providing compute and azure credits. Happy Buzaaba is supported by the Center for Digital Humanities, the Africa Humanities Colloquium, and the Africa World Initiative at Princeton.

\bibliography{custom}

\appendix
\begin{table*}[t]
    \centering
    \small
    
\resizebox{\textwidth}{!}{%
\setlength{\tabcolsep}{2pt}
\setlength{\defaultaddspace}{5pt}
\setlength{\aboverulesep}{0pt}
\setlength{\belowrulesep}{0pt}

\begin{tabular}{@{\hspace{5pt}}lcccccccccccccccccccc@{\hspace{5pt}}}
\toprule
Model & Size & \textbf{eng} & \textbf{\textit{fra}} & \textbf{amh} & \textbf{hau} & \textbf{ibo} & \textbf{kin} & \textbf{orm} & \textbf{sna} & \textbf{sot} & \textbf{swa} & \textbf{xho} & \textbf{yor} & \textbf{zul} & \textbf{\textit{ewe}} & \textbf{\textit{lin}} & \textbf{\textit{lug}} & \textbf{\textit{twi}} & \textbf{\textit{wol}} & \textbf{Avg}$^\dagger$ \\
\midrule
& \multicolumn{19}{c}{\tt AfriXNLI (English Prompt)} \\
\midrule

Llama-3.1 & 8B & \cc{87bc9f} 59.2 & \cc{93c2a8} 55.7 & \cc{c5ddcf} 41.2 & \cc{c4dccf} 41.3 & \cc{c1dbcc} 42.2 & \cc{dbe8e1} 34.7 & \cc{d8e7df} 35.3 & \cc{d2e3d9} 37.3 & \cc{d9e7df} 35.2 & \cc{c2dbcd} 41.8 & \cc{d4e5dc} 36.5 & \cc{c5ddcf} 41.2 & \cc{cee2d7} 38.3 & \cc{dce9e1} 34.3 & \cc{e6eee9} 31.3 & \cc{d6e6dd} 36.2 & \cc{d3e4db} 36.8 & \cc{dfeae4} 33.5 & \cc{d2e3d9} 37.3 \\
\addlinespace
Lugha-Llama & 8B & \cc{87bc9f} 59.0 & \cc{91c2a7} 56.0 & \cc{b8d6c5} 45.0 & \cc{bcd8c8} 43.7 & \cc{bad7c7} 44.3 & \cc{c8ded2} 40.2 & \cc{c2dbcd} 42.0 & \cc{b9d6c6} 44.5 & \cc{c2dbcd} 41.8 & \cc{bcd8c8} 43.7 & \cc{b2d3c1} 46.5 & \cc{bfdacb} 42.8 & \cc{b2d3c0} 46.7 & \cc{dfebe4} 33.3 & \cc{deeae3} 33.8 & \cc{cee2d7} 38.3 & \cc{e2ece6} 32.7 & \cc{e2ece6} 32.7 & \cc{c6ddd0} 40.8 \\
 $\;\;$-edu & 8B & \cc{84ba9c} 60.0 & \cc{8fc1a6} 56.7 & \cc{b5d4c3} 45.7 & \cc{bbd7c8} 44.0 & \cc{bad7c7} 44.3 & \cc{cde1d6} 38.7 & \cc{c0dacc} 42.3 & \cc{b8d6c5} 44.8 & \cc{bcd8c8} 43.7 & \cc{c0dacc} 42.3 & \cc{b7d5c4} 45.2 & \cc{bed9ca} 43.0 & \cc{bad7c7} 44.3 & \cc{ddeae3} 34.0 & \cc{e2ece6} 32.5 & \cc{cce1d5} 39.0 & \cc{dfeae4} 33.5 & \cc{e2ece6} 32.5 & \cc{c6ddd1} 40.6 \\
 $\;\;$-math & 8B & \cc{83ba9c} 60.2 & \cc{89bda1} 58.3 & \cc{b8d6c5} 44.8 & \cc{b2d3c1} 46.5 & \cc{b0d2bf} 47.2 & \cc{c2dbcd} 41.8 & \cc{b8d6c5} 44.8 & \cc{b0d2bf} 47.2 & \cc{b8d6c5} 45.0 & \cc{aed1bd} 47.8 & \cc{aed1bd} 47.8 & \cc{b7d5c4} 45.2 & \cc{bad7c7} 44.3 & \cc{d7e6de} 35.7 & \cc{dfeae4} 33.5 & \cc{cce0d5} 39.2 & \cc{dfeae4} 33.5 & \cc{e4ede8} 32.0 & \cc{c1dbcc} 42.3 \\
\addlinespace
\textbf{aya-101} & 13B & \cc{77b492} \bf 63.8 & \cc{80b899} \bf 61.2 & \cc{91c1a7} \bf 56.2 & \cc{93c3a9} \bf 55.5 & \cc{9bc6ae} \bf 53.3 & \cc{91c1a7} \bf 56.2 & \cc{9ec8b1} \bf 52.3 & \cc{8ec0a4} \bf 57.0 & \cc{97c4ab} \bf 54.5 & \cc{98c5ac} \bf 54.2 & \cc{8dbfa4} \bf 57.2 & \cc{a4ccb6} \bf 50.5 & \cc{99c6ad} \bf 53.8 & \cc{b7d5c4} \bf 45.2 & \cc{e1ebe5} 33.0 & \cc{9bc6ae} \bf 53.3 & \cc{a7cdb8} \bf 49.7 & \cc{d6e6dd} \bf 36.0 & \cc{a2cbb5} \bf 51.1 \\
LLaMAX3 & 8B & \cc{97c4ab} 54.5 & \cc{9fc9b1} 52.2 & \cc{b6d5c4} 45.3 & \cc{b3d3c1} 46.3 & \cc{b9d6c6} 44.5 & \cc{d3e4db} 36.8 & \cc{c5ddcf} 41.2 & \cc{b7d5c4} 45.2 & \cc{c8ded2} 40.2 & \cc{b1d2c0} 46.8 & \cc{b7d5c4} 45.2 & \cc{bdd9ca} 43.2 & \cc{bad7c7} 44.2 & \cc{dfeae4} 33.5 & \cc{dae8e0} \bf 34.8 & \cc{c8ded2} 40.2 & \cc{dce9e1} 34.3 & \cc{d8e7df} 35.3 & \cc{c5ddcf} 41.1 \\
AfroLlama-V1 & 8B & \cc{abcfbb} 48.5 & \cc{bbd7c8} 44.0 & \cc{dfebe4} 33.3 & \cc{b8d6c5} 44.8 & \cc{dae8e0} 34.8 & \cc{d8e7df} 35.3 & \cc{dde9e2} 34.2 & \cc{dce9e1} 34.3 & \cc{dbe8e1} 34.7 & \cc{b5d4c3} 45.8 & \cc{b9d6c6} 44.5 & \cc{c0dacc} 42.3 & \cc{b9d6c6} 44.7 & \cc{e1ebe5} 33.0 & \cc{e1ebe5} 33.0 & \cc{d9e7df} 35.2 & \cc{d3e4db} 36.8 & \cc{dbe9e1} 34.5 & \cc{d1e3d9} 37.6 \\
AfriInstruct & 7B & \cc{aacfbb} 48.8 & \cc{accfbc} 48.3 & \cc{d5e5dc} 36.3 & \cc{dbe9e1} 34.5 & \cc{d1e3d9} 37.5 & \cc{d8e7df} 35.3 & \cc{dde9e2} 34.2 & \cc{d2e3d9} 37.3 & \cc{d3e4da} 37.2 & \cc{d8e7de} 35.5 & \cc{d1e3d9} 37.5 & \cc{d3e4da} 37.2 & \cc{cce1d5} 39.0 & \cc{dfebe4} 33.3 & \cc{e3ece7} 32.3 & \cc{e1ece6} 32.8 & \cc{edf2ef} 29.3 & \cc{d7e6de} 35.7 & \cc{d9e7df} 35.3 \\
InkubaLM & 0.4B & \cc{deeae3} 33.7 & \cc{e0ebe5} 33.2 & \cc{dde9e2} 34.2 & \cc{ddeae3} 34.0 & \cc{ddeae3} 34.0 & \cc{e2ece6} 32.5 & \cc{e2ece6} 32.5 & \cc{e4ede8} 32.0 & \cc{e5ede8} 31.8 & \cc{deeae3} 33.8 & \cc{dfebe4} 33.3 & \cc{deeae3} 33.7 & \cc{e0ebe5} 33.2 & \cc{deeae3} 33.7 & \cc{e1ebe5} 33.0 & \cc{e7efea} 31.2 & \cc{e4ede8} 32.2 & \cc{e5ede8} 31.8 & \cc{e1ebe5} 32.9 \\
\bottomrule
\end{tabular}
}
    \caption{Results of running AfriXNLI with an English prompt template, instead of a language-specific prompt template.}
    \label{tab:afrixnli_en_direct}
\end{table*}

\begin{table*}[t]
    \centering
     \scriptsize 
    
\resizebox{0.8\textwidth}{!}{%
\setlength{\tabcolsep}{2pt}
\setlength{\defaultaddspace}{5pt}
\setlength{\aboverulesep}{0pt}
\setlength{\belowrulesep}{0pt}

\begin{tabular}{@{\hspace{5pt}}lccccccccccc@{\hspace{5pt}}}
\toprule
Model & Size & \textbf{ibo} & \textbf{swa} & \textbf{hau} & \textbf{kin} & \textbf{zul} & \textbf{yor} & \textbf{\textit{bem}} & \textbf{\textit{fon}} & \textbf{\textit{twi}} & 
\textbf{Avg}$^\dagger$ \\
\midrule
& \multicolumn{10}{c}{\tt AfriQA} \\
\midrule

Llama-3.1 & 8B & \cc{87bc9f} 44.1 & \cc{93c2a8} 42.3 & \cc{c5ddcf} 19.5 & \cc{c4dccf} 15.5 & \cc{c1dbcc} 12.5 & \cc{dbe8e1} 13.0 & \cc{d8e7df} 11.4 & \cc{d2e3d9} 12.6 & \cc{d9e7df} 9.9 & \cc{d2e3d9} 20.1\\
\addlinespace
Lugha-Llama & 8B & \cc{87bc9f} 57.0 & \cc{91c2a7} 47.5 & \cc{b8d6c5} 49.2 & \cc{bcd8c8} 44.9 & \cc{bad7c7} 37.7 & \cc{c8ded2} 37.3 & \cc{c2dbcd} 16.8 & \cc{b9d6c6} 10.3 & \cc{c2dbcd} 6.9 & \cc{c6ddd0} 34.2
\\
 $\;\;$-edu & 8B & \cc{84ba9c} 57.7 & \cc{8fc1a6} 43.1 & \cc{b5d4c3} 39.8 & \cc{bbd7c8} 36.8 & \cc{bad7c7} 29.3 & \cc{cde1d6} 33.9 & \cc{c0dacc} 11.6 & \cc{b8d6c5} 11.2 & \cc{bcd8c8} 9.2 & \cc{c6ddd1} 30.3 
 \\ 
 $\;\;$-math & 8B & \cc{83ba9c} 66.1 & \cc{89bda1} 52.8 & \cc{b8d6c5} 48.2 & \cc{b2d3c1} 45.0 & \cc{b0d2bf} 45.9 & \cc{c2dbcd} 41.1 & \cc{b8d6c5} 16.9 & \cc{b0d2bf} 12.4 & \cc{b8d6c5} 10.8 & \cc{c1dbcc} 37.7 
 \\ 
\addlinespace
aya-101 & 13B & \cc{77b492} \bf 85.0 & \cc{80b899} \bf 70.8 & \cc{91c1a7} \bf 73.8 & \cc{93c3a9} \bf 68.7 & \cc{9bc6ae} \bf 79.5 & \cc{91c1a7} \bf 68.8 & \cc{9ec8b1} \bf 43.1 & \cc{8ec0a4} \bf 30.6 & \cc{97c4ab} \bf 42.7 & \cc{a2cbb5} \bf 62.6 
\\
LLaMAX3 & 8B & \cc{abcfbb} 2.27 & \cc{bbd7c8} 1.83 & \cc{dfebe4} 1.67 & \cc{b8d6c5} 2.56 & \cc{dae8e0} 1.55 & \cc{d8e7df} 1.57 & \cc{dde9e2} 1.23  & \cc{dce9e1} 3.64  & \cc{dbe8e1} 1.26 & \cc{d1e3d9} 1.97
\\
AfroLlama-V1 & 8B & \cc{abcfbb} 10.3 & \cc{bbd7c8} 31.6 & \cc{dfebe4} 33.0 & \cc{b8d6c5} 12.2 & \cc{dae8e0} 27.6 & \cc{d8e7df} 28.0 & \cc{dde9e2} 11.6 & \cc{dce9e1} 8.2 & \cc{dbe8e1} 8.4 & \cc{d1e3d9} 19.0 
\\
AfriInstruct & 7B & \cc{aacfbb} 35.3 & \cc{accfbc} 32.0 & \cc{d5e5dc} 29.6 & \cc{dbe9e1} 25.6 & \cc{d1e3d9} 28.7 & \cc{d8e7df} 30.5 & \cc{dde9e2} 5.7 & \cc{d2e3d9} 5.8 & \cc{d3e4da} 3.8 & \cc{d9e7df} 21.9 
\\
InkubaLM & 0.4B & \cc{aacfbb} 1.1 & \cc{accfbc} 1.3 & \cc{d5e5dc} 1.2 & \cc{dbe9e1} 2.7 & \cc{d1e3d9} 2.2 & \cc{d8e7df} 1.3 & \cc{dde9e2} 1.9 & \cc{d2e3d9} 2.0 & \cc{d3e4da} 1.3 & \cc{d9e7df} 1.7 
\\
\bottomrule
\end{tabular}
}
    \caption{Results of Lugha-Llama models and baselines on AfriQA~\cite{ogundepo-etal-2023-cross}.}
    \label{tab:AfriQA-results}
\end{table*}
\section{Appendix}

\subsection{Baselines}
\label{app:baselines}
\textbf{Aya-101} is based on mT5~\citep{xue2020mt5} backbone and has been instruction tuned on diverse multilingual prompted task datasets covering 101 languages including ten African languages in IrokoBench. Llama-3-8B was pre-trained from scratch on 15 trillion tokens of which 8\% are non-English, and was further adapted to \textbf{Llama-3.1-8B} \citep{dubey2024llama}.
\textbf{LLaMaX3-8B}~\citep{lu2024llamax} is based on continual pre-training of LLaMa-3-8B on massive multilingual data of 101 languages. \textbf{AfroLlama-v1} is based on LORA-based continual pre-training and instruction tuning of Llama 3 8B on six languages of Africa: Swahili, Xhosa, Zulu, Yoruba, Hausa and English. \textbf{InkubaLM-0.4B}~\citep{tonja2024inkubalm} has been pre-trained from scratch on five African languages: isiXhosa, isiZulu, Swahili, Hausa and Yoruba. 
The closest model to our pre-training setup is \textbf{AfriInstruct}~\citep{uemura_afriinstruct}---based on continual pre-training of Llama 2 7B on WURA dataset followed by instruction-tuning on several African languages tasks. 

\subsection{Additional Results}
\label{app:results}

\begin{table}[h]
    \centering
    \small
    \resizebox{\linewidth}{!}{
    \begin{tabular}{lccc}
        \toprule
        & \tt AfriMMLU & \tt AfriXNLI \\
        \midrule
        Llama-3.2-3B & 25.59 & 32.61 \\
        \midrule
        WURA only & 26.05 & \bf 33.88 \\
        $\;\;$ w/o UniMax & 26.00 & 33.11 \\
        \cmidrule(lr){1-1}
        80\% WURA : 20\% FineWeb-Edu & 26.66 & 33.57 \\
        \cmidrule(lr){1-1}
        60\% WURA : 40\% FineWeb-Edu & \bf 26.70 & 33.55 \\
        $\;\;$ w/o UniMax & 26.44 & 32.82 \\
        \bottomrule
        \end{tabular}
    }
    \caption{We ablate the importance of UniMax sampling \cite{chung2023unimax} by continuning to train Llama-3.2-3B by 5B tokens. We report the same average over non-high-resource languages as in \autoref{tab:main_results}.}
    \label{tab:ablations}
\end{table}

\paragraph{Ablations.}
We explore some design decisions by running small-scale ablations by continuing to train Llama-3.2-3B for 1200 steps (5B tokens).
\autoref{tab:ablations} shows the
results of comparing UniMax sampling vs. sampling languages according to their natural frequency in the WURA corpus.
The trends of our main results still hold true at the 3B-parameter scale, e.g., adding FineWeb-edu to WURA helps on AfriMMLU. However, the magnitude of the gain is smaller compared to the 8B models.
\autoref{tab:ablations} also shows that adding 40\% FineWeb-edu data performs marginally better than only adding 20\%.

\paragraph{Prompt formats.}
The AfriXNLI also has an English prompt template in the IrokoBench dataset \citep{adelani2024irokobench}.
We opt for choosing the native-language prompt template as our main evaluation, since we believe that it better reflects the actual use cases of these models.
However, we feature the English prompt in \autoref{tab:afrixnli_en_direct}.
We find that in this setting, the aya-101 model performs best. We note that it even outperforms Llama-3.1-8B on the English examples, despite Llama-3.1-8B's strong overall benchmark performance. Similarly, in \autoref{tab:AfriQA-results} aya-101 model performs best on AfriQA \cite{ogundepo-etal-2023-cross} compared to baselines.

\end{document}